\documentclass[conference]{IEEEtran}
\usepackage{stmaryrd}
\usepackage{amsfonts}

\usepackage{graphicx}
\usepackage{tabularx}
\usepackage{colortbl}
\usepackage{setspace}

\usepackage{url}
\usepackage{amsmath}
\usepackage{amssymb}
\usepackage{array}
\usepackage{times}
\usepackage{epsfig}
\usepackage{subfigure}

\graphicspath{{./imgs/}}

\IEEEoverridecommandlockouts    

\begin{document}

\title{\LARGE\bf Discriminative $k$-Means Clustering}

\author{Ognjen Arandjelovi\'c\\Centre for Pattern Recognition and Data Analytics\\Deakin University, Australia\\\texttt{ognjen.arandjelovic@gmail.com}}


\maketitle

\begin{abstract}
The $k$-means algorithm is a partitional clustering method. Over 60 years old, it has been successfully used for a variety of problems. The popularity of $k$-means is in large part a consequence of its simplicity and efficiency. In this paper we are inspired by these appealing properties of $k$-means in the development of a clustering algorithm which accepts the notion of ``positively'' and ``negatively'' labelled data. The goal is to discover the cluster structure of both positive and negative data in a manner which allows for the discrimination between the two sets. The usefulness of this idea is demonstrated practically on the problem of face recognition, where the task of learning the scope of a person's appearance should be done in a manner which allows this face to be differentiated from others.
\end{abstract}


\section{Introduction}
\PARstart{I}{n} data analysis, clustering refers to the process of discovering groups (clusters) of similar data points. Being a vital tool in the exploration, sparsification and dimensionality reduction of data, it is unsurprising to observe that clustering is intensively used in a wide range of applications, from protein sequencing~\cite{VoevBalcRoglTeng+2012} to astronomical surveys of the sky~\cite{JangHend2007}. The use of clustering is even more ubiquitous in image and multimedia processing: at the lowest level, clustering is used to build vocabularies of low-level features used to represent individual objects~\cite{Lowe2004,AranZiss2011,Aran2012f}, images~\cite{JegoDouzSchmPere2010} or elementary motions~\cite{Aran2011a}; at medium-level, clustering may be used to list the cast of a movie~\cite{AranCipo2006c} or detect coherently moving objects~\cite{OchsBrox2012,MartAran2010}; at the highest level, video clips or images may themselves be clustered by the similarity of their content~\cite{Lian2013,GirgShipWilc2011}.

Despite continuing major research efforts, clustering remains a most challenging problem. The application-specific notion of what a good cluster is, the volume and dimensionality of data, its sparsity and structure, and the number of clusters, are only some of the factors which govern one's choice of the clustering methodology. Popular algorithms include those based on spectral approaches~\cite{NgJordWeis2001}, various non-parametric formulations e.g.\ using the Dirichlet process~\cite{McCuYang2008}, information theoretic ideas~\cite{RobeHolmDeni2001}, and many others~\cite{DudaHartStor2001}.

Yet, notwithstanding the cornucopia of sophisticated clustering models described in the literature, to this day the most popular and widely used clustering approach remains to be the simple $k$-means algorithm~\cite{Jain2010}.

\section{$K$-means clustering}\label{s:kmeans}
Let $X = \{ x_1, x_2, \ldots, x_n \}$ be a set of $d$-dimensional points. The $k$-means algorithm partitions the points into $K$ clusters, $X_1, \ldots, X_K$, so that each datum belongs to one and only one cluster. In addition, an attempt is made to minimize the sum of squared distances between each data point and the empirical mean of the corresponding cluster. In other words, the $k$-means algorithm attempts to minimize the following objective function:
\begin{align}
  J(X_1,\ldots,X_k) = \sum_{i=1}^k \sum_{x \in X_i} \|c_i-x\|^2,
  \label{e:kmeansObj}
\end{align}
where the empirical cluster means are calculated as:
\begin{align}
  c_i = \sum_{x \in X_i} x ~/~ \big| X_i \big|.
\end{align}
The exact minimization of the objective function in Eq.~\eqref{e:kmeansObj} is an NP-hard problem~\cite{Dasg2007}. Instead, the $k$-means algorithm only guarantees convergence to a local minimum. Starting from an initial guess, the algorithm iteratively updates cluster centres and data-cluster assignments until (i) a local minimum is attained, or (ii) an alternative stopping criterion is met (e.g.\ the maximal desired number of iterations or a sufficiently small sum of squared distances).

The $k$-means algorithm starts from an initial guess of cluster centres $c_1^{(0)},c_2^{(0)},\ldots,c_k^{(0)}$. Often, this is achieved simply by choosing $k$ data points at random as the centres of the initial clusters, although more sophisticated initialization methods have been proposed~\cite{PenaLozaLarr1999,KhanAhma2004}. Then, at each iteration $t=0,\ldots$ the new datum-cluster assignment is computed:
\begin{align}
  X_i^{(t)} = \{ x~:~ x \in X \wedge \arg \min_j \big\|x - c_j^{(t)} \big\|^2 = i \}.
  \label{e:kmeansUpd1}
\end{align}
In other words, each datum is assigned to the cluster with the nearest (in Euclidean sense) empirical mean. Lastly, the locations of cluster centres are re-computed from the new assignments by finding the mean of the data assigned to each cluster:
\begin{align}
  c_i^{(t+1)} = \sum_{x \in X_i^{(t)}} x ~/~ \big| X_i^{(t)} \big|.
  \label{e:kmeansUpd2}
\end{align}
The algorithm is guaranteed to converge because neither of the updates in Eq.~\eqref{e:kmeansUpd1} nor Eq.~\eqref{e:kmeansUpd2} can ever increase the objective function of Eq.~\eqref{e:kmeansObj}.

Various extensions of the original $k$-means algorithm include fuzzy $c$-means \cite{Dunn1973}, $k$-medoid \cite{KaufRous2005}, kernel $k$-means \cite{SchoSmolMull1998}. Their relevance to the topic of the present paper is tangential and the interested reader is referred to the cited publications for further information.

\section{Extended $K$-means clustering}\label{s:ekmeans}
In the present paper we are interested in the problem of clustering data which comprises two types of data points, which we will call ``positively labelled'' and ``negatively labelled'', and which should be clustered in such a way to allow for the two types to be discriminated between. Our goal is to inherit the simple ideas underlying the classical $k$-means algorithm, while extending the algorithm to deal with this novel complexity in the data.

\begin{figure*}[tb]
  \centering
  \subfigure[]{\includegraphics[width=0.75\textwidth]{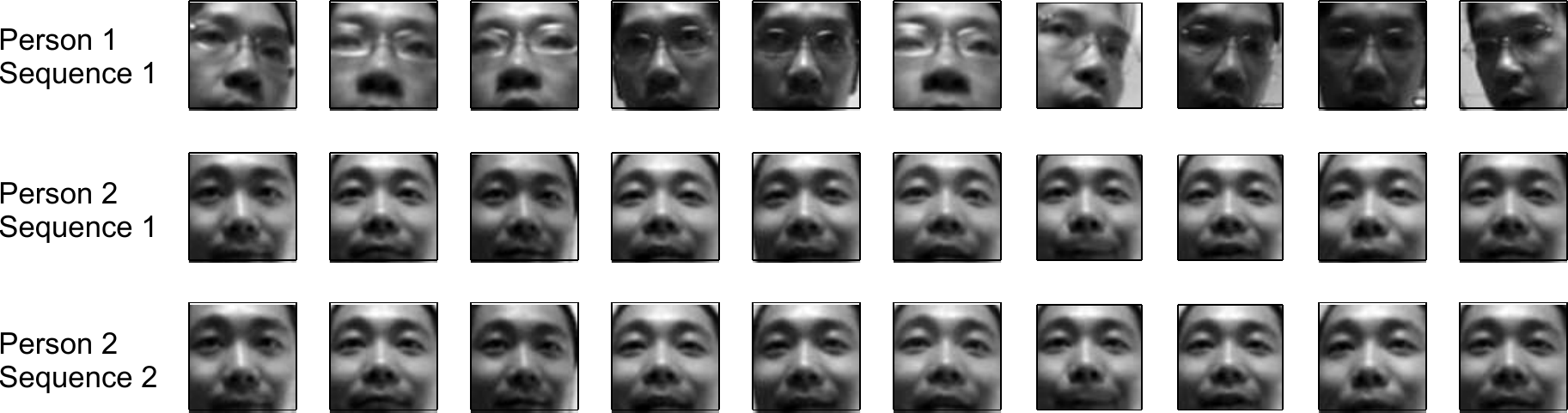}}
  \subfigure[]{\includegraphics[width=0.33\textwidth]{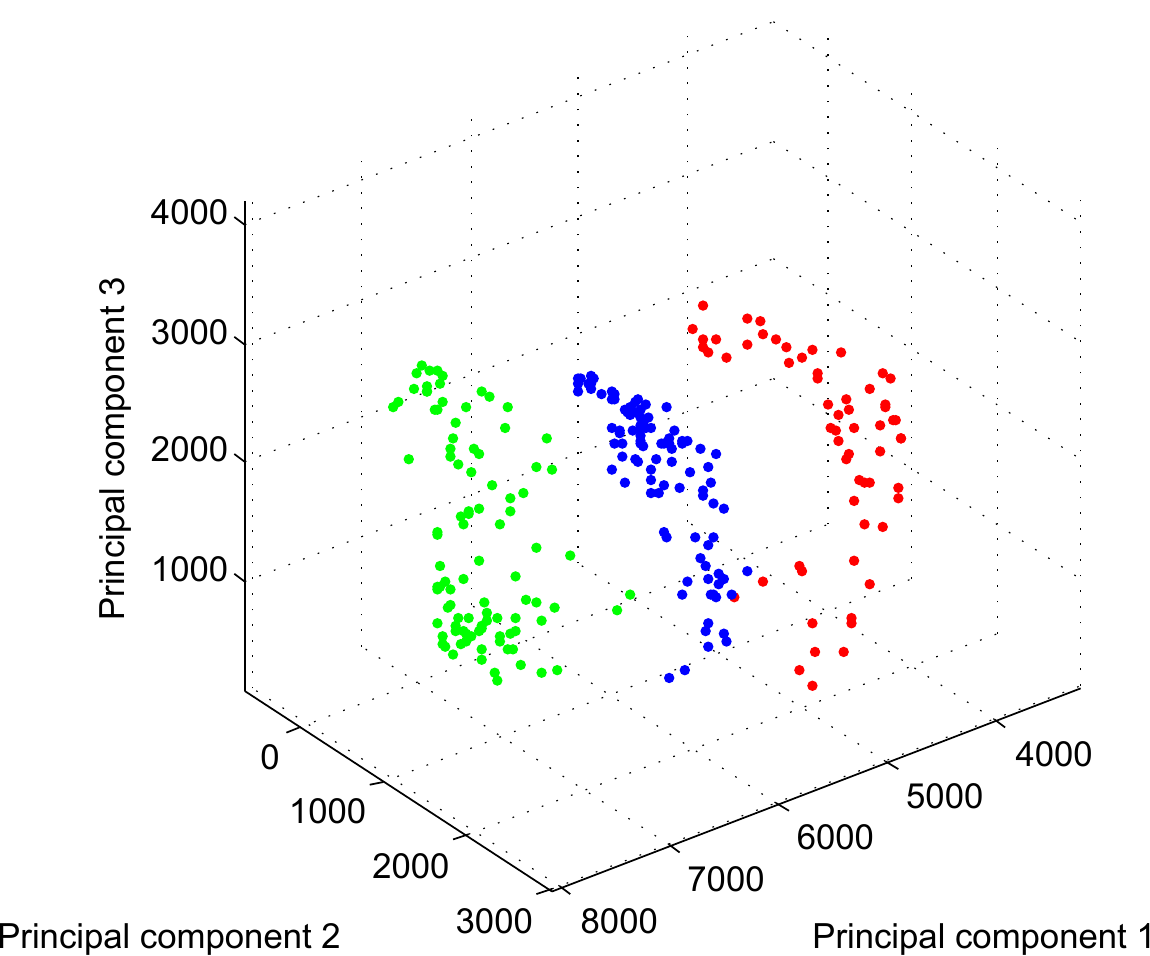}}\hspace{60pt}
  \subfigure[]{\includegraphics[width=0.33\textwidth]{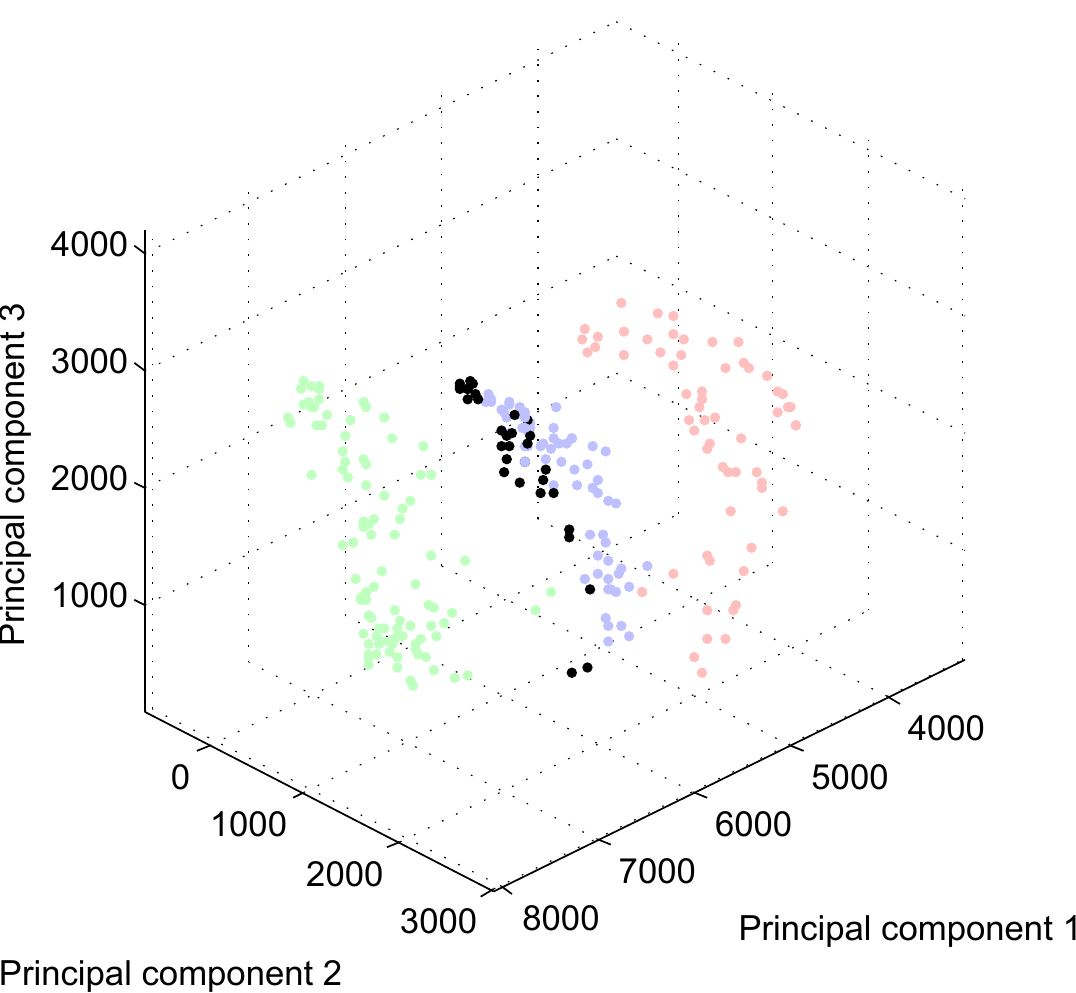}}
  \caption{  (a) Three 10s long face motion sequences (due to space constraints only every 10th detected face is shown). (b) The three sets of rasterized face images shown projected to the first three principal components. (c) Misclassified faces of the blue set (32\% misclassification rate) when the green and red sets are used for training and the $k$-means algorithm is used to represent the corresponding manifold structures. }
  \label{f:faces}
\end{figure*}

\begin{figure}[tb]
  \centering
  \footnotesize
  \includegraphics[width=0.45\textwidth]{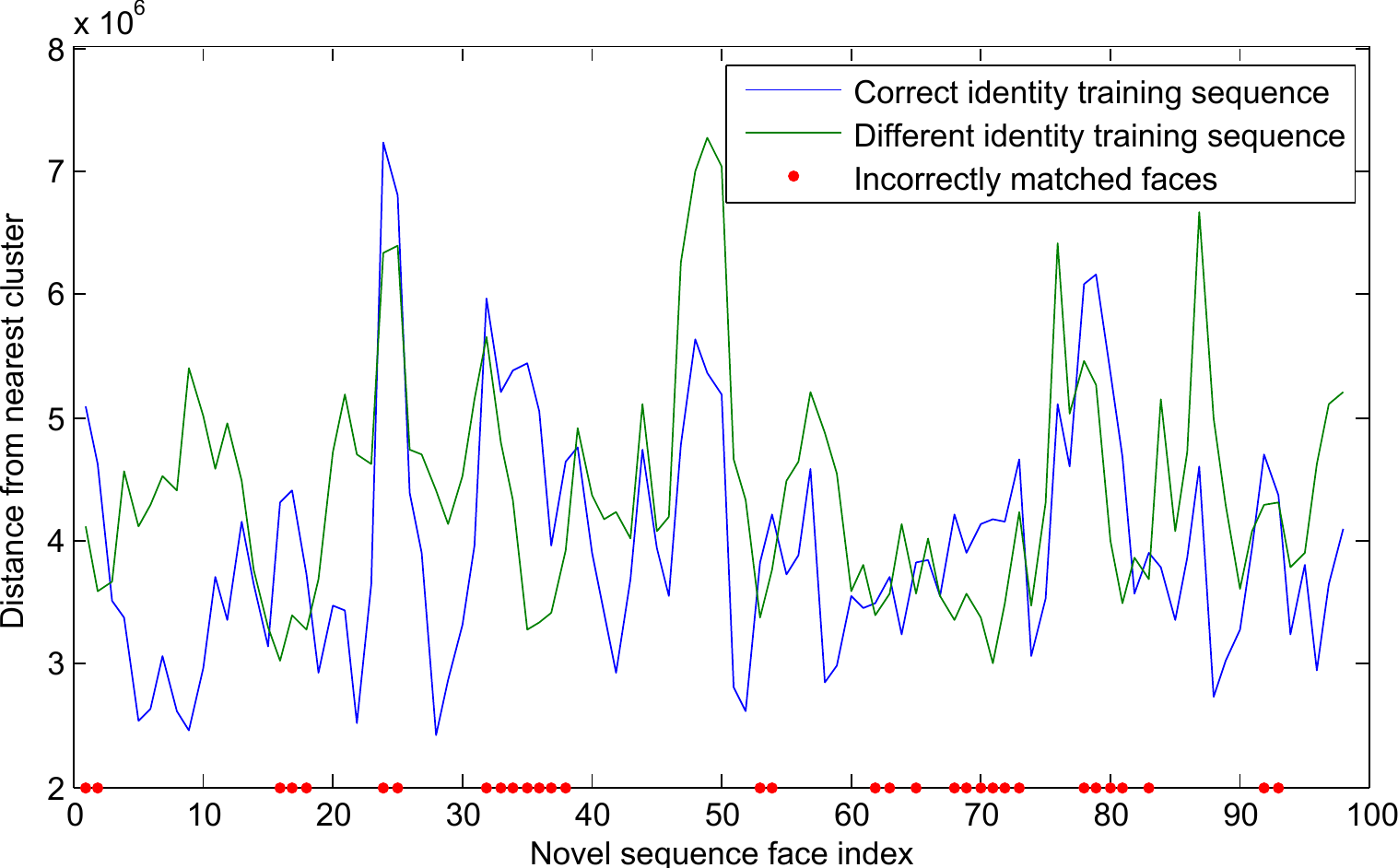}
  \caption{ Distances of the faces from the novel video sequence to from the nearest clusters obtained using conventional $k$-means clustering performed on the
            two training image sequences. The blue and green lines show the distance from the nearest cluster of the training sequence respectively correctly
            and incorrectly matched by the identity. Red dots mark the incorrectly classified faces. Approximately 32\% of the novel data is misclassified. }
  \label{f:facedists}
\end{figure}

To motivate this problem formulation using a concrete practical example, consider Figure~\ref{f:faces}(a). Its top row shows 10 detected faces from a 10s motion sequence of a person. The other two rows similarly show two sets of 10 detected faces from motion sequences of the same length, but of a different person. The two sequences of the second person were acquired at two different times. Figure~\ref{f:faces}(b) shows the corresponding face sets (approximately 100 faces per sequence), rasterized and projected onto the first three principal linear components. The green set corresponds to the sequence of Person~1, while the blue and red sets correspond to the two sequences of Person~2. We wish to sparsify and discover the structure of the sets corresponding to different frequencies; at the same time we wish to exploit the knowledge that some of the sequences belong to different people, which we may wish to distinguish between (i.e.\ we may wish to retain discriminative information for the purposes of recognition). Consider the scenario in which only the sequences corresponding to the green and red sets are available for training, the $k$-means algorithm is applied independently to both (we used $k=5$), and then the obtained cluster centres used to classify each of the novel blue faces. Figure~\ref{f:faces}(c) shows the same data as Figure~\ref{f:faces}(b), with the exception that the misclassified blue faces are emphasized. Specifically, in this case a third of novel faces are incorrectly recognized. The plot in Figure~\ref{f:facedists} shows this even more convincingly, by plotting the datum-by-datum distance from the nearest cluster of each of the training sequences. As we will show in Section~\ref{s:eval}, a major reason for this poor performance is to be found in the lack of available class information in clustering. Indeed, when the proposed algorithm is applied, no misclassification occurs in this instance.

\begin{figure}[htb]
  \centering
  \vspace{0pt}
  \subfigure[Conventional $k$-means]{\includegraphics[trim=0 -10pt 0 0,clip,width=0.35\textwidth]{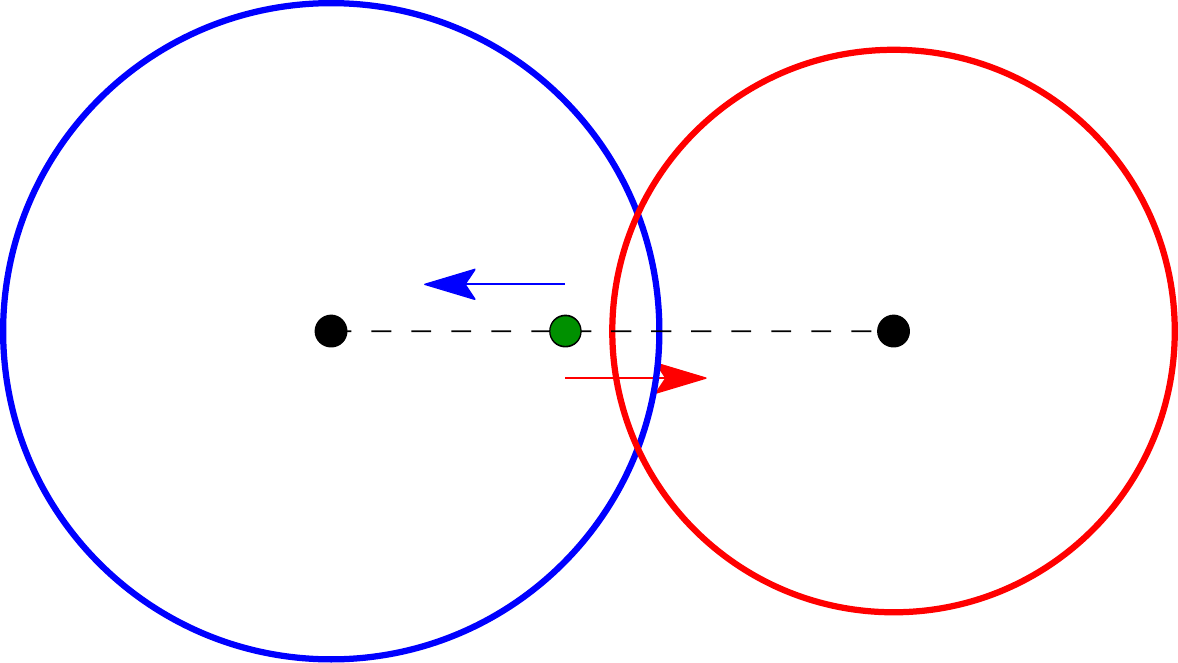}}\vspace{10pt}
  \subfigure[Proposed discriminative $k$-means]{\includegraphics[trim=0 -10pt 0 0,clip,width=0.35\textwidth]{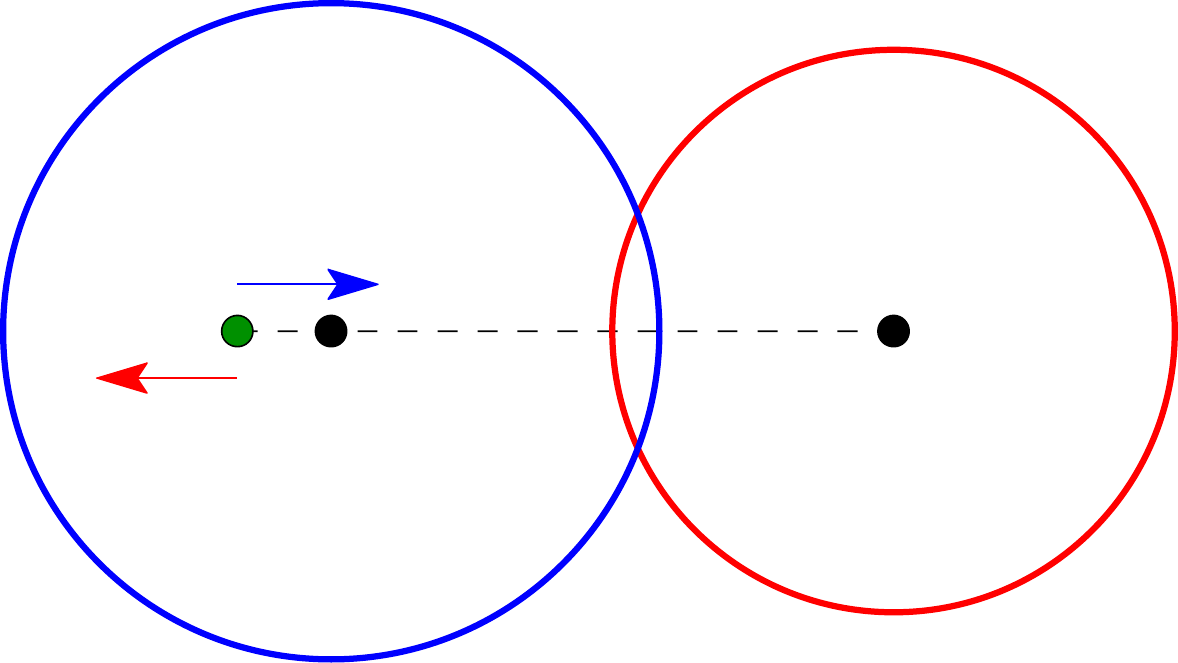}}
  \caption{ The effect of different groups of data on cluster centres in (a) the $k$-means algorithm, and (b) the proposed algorithm when the groups correspond to different class labels. }
  \label{f:mean}
\end{figure}

Our key idea is motivated by the second $k$-means update equation, i.e.\ Eq.~\ref{e:kmeansUpd2}. Instead of considering its `batch' form introduced in the previous section, consider the form that the update takes if we partition the data points belonging to the $i$-th cluster into two; call these clusters $\dot{X}_i$ and $\ddot{X}_i$, where the iteration superscript has been temporarily omitted for clarity. Let the corresponding empirical means be $\dot{c}_i = \sum_{x \in \dot{X}_i} x ~/~ \big| \dot{X}_i \big|$ and $\ddot{c}_i = \sum_{x \in \ddot{X}_i} x ~/~ \big| \ddot{X}_i \big|$. The computation of the empirical mean of the entire cluster $X_i$ can then be seen as taking $\dot{c}_i$ as the initial estimate and adjusting it by adding to it a correction component in the direction $(\ddot{c}_i - \dot{c}_i)$:
\begin{align}
  c_i^{(t+1)} = \dot{c}_i^{(t+1)} + w \times (\ddot{c}_i^{(t+1)}- \dot{c}_i^{(t+1)}).
  \label{e:kmeansUpd2new}
\end{align}
The parameter $w$ determines the magnitude of the correction in the direction $(\ddot{c}_i - \dot{c}_i)$ and it is dependent on the number of data points in each of the partitions $\dot{X}_i$ and $\ddot{X}_i$:
\begin{align}
  w = \frac { \big| \ddot{X}_i \big| } {\big| \dot{X}_i \big| + \big| \ddot{X}_i \big|}
  \label{e:weight}
\end{align}
This can be seen as the data points in $\ddot{X}_i$ bringing new evidence for the centre of the corresponding cluster, pulling it towards its centre, as illustrated in Figure~\ref{f:mean}(a).

However, returning to the case of positively and negatively labelled data, what if the two sets of data points, $\dot{X}_i$ and $\ddot{X}_i$ correspond to data which are differently labelled?  What should the effect on the cluster centre be if data in $\dot{X}_i$ is positively labelled and data in $\ddot{X}_i$ negatively? Motivated by the formulation expressed by Eq.~\ref{e:kmeansUpd2new}, in the paper we propose that two things should happen. Firstly, the cluster should split into two. This is because we know that we do not want to have clusters which contain data points with both types of labels. Secondly, we propose that the effect of each of the new children clusters should be that of \emph{repelling} (rather than attracting like in Eq.~\ref{e:kmeansUpd2new}) the other, as illustrated in Figure~\ref{f:mean}(b).

Formally, our algorithm can be summarized by the following sequence of steps:
\begin{list}{\labelitemi}{\leftmargin=2.5em}
  \item[Initialization:] Initially, all data belongs to a single cluster.\\
  \item[Iteration $t$:] As in the $k$-means algoritithm, each data point is assigned to its nearest cluster:
        \begin{align}
          X_i^{(t)} = \{ x~:~ \arg \min_j \big\|x - c_j^{(t)} \big\|^2 = i \}.
        \end{align}
  \item[Iteration $t$:] If a cluster $X_i^{(t)}$ contains both positively and negatively labelled data, split it into two. The centres of the two child clusters become:
        \begin{align}
          c_i^{(t+1)}     &= \dot{c}_i^{(t+1)} - w \times (\ddot{c}_i^{(t+1)}- \dot{c}_i^{(t+1)}), \text{ and}\\
          c_{n+1}^{(t+1)} &= \ddot{c}_i^{(t+1)} - w \times (\dot{c}_i^{(t+1)}- \ddot{c}_i^{(t+1)}),
        \end{align}
        where $n$ is the previous number of clusters, $\dot{c}_i^{(t+1)}$ is the mean of positively labelled data in $X_i^{(t)}$ and $\ddot{c}_i^{(t+1)}$ is the mean of negatively labelled data in $X_i^{(t)}$, and $w$ a non-negative weight.\\
  \item[Iteration $t$:] If a cluster $X_i^{(t)}$ contains only positively or negatively labelled data, update its empirical mean as in the $k$-means algorithm:
        \begin{align}
          c_i^{(t+1)} = \sum_{x \in X_i^{(t)}} x ~/~ \big| X_i^{(t)} \big|.
       \end{align}
  \item[Termination:] Terminate when the maximal desired number of clusters is reached or when the algorithm has converged (guaranteed in a similar manner to the conventional $k$-means algorithm.
\end{list}

\section{Evaluation}\label{s:eval}

\subsection{Synthetic data}
\begin{figure*}[tb]
  \centering
  \footnotesize
  \subfigure[Experiment 1: Synthetic data set 1]{\includegraphics[trim=0 -10pt 0 0,clip,width=1\textwidth]{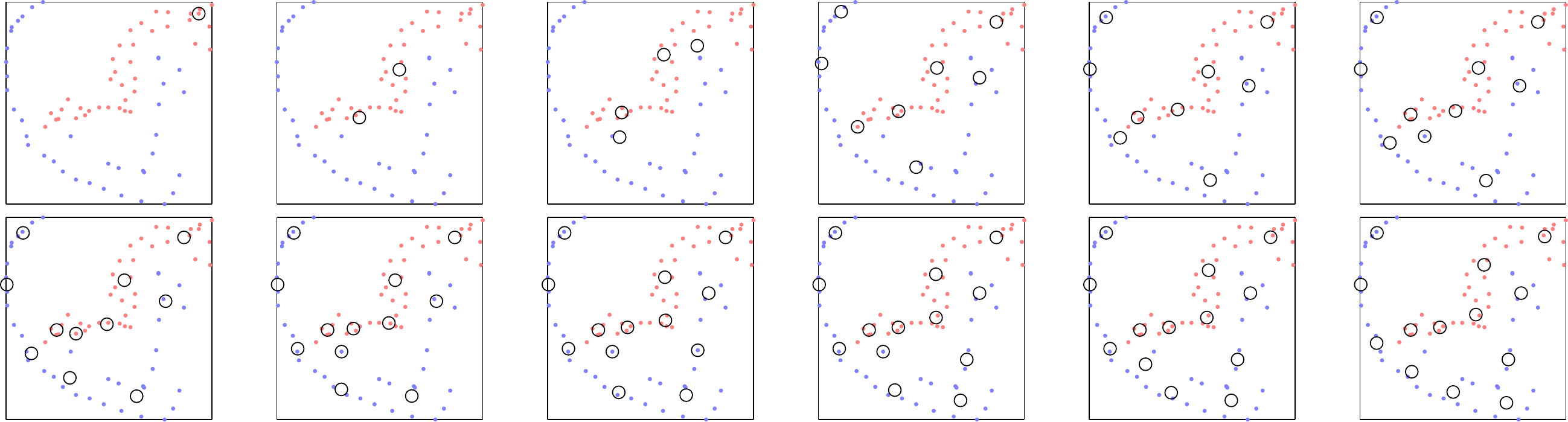}}\vspace{12pt}
  \subfigure[Experiment 2: Synthetic data set 2 (expanded synthetic data set 1)]{\includegraphics[trim=0 -10pt 0 0,clip,width=1\textwidth]{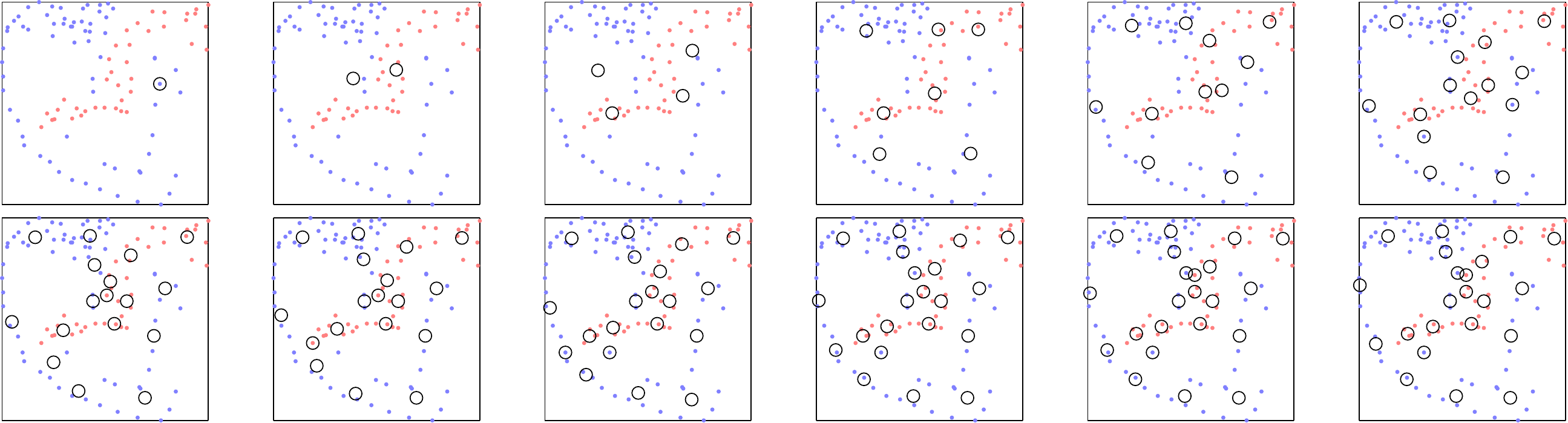}}\vspace{12pt}
  \subfigure[Experiment 3: Synthetic data set 2, algorithm run from the final state of Experiment 1]{\includegraphics[trim=0 -10pt 0 0,clip,width=1\textwidth]{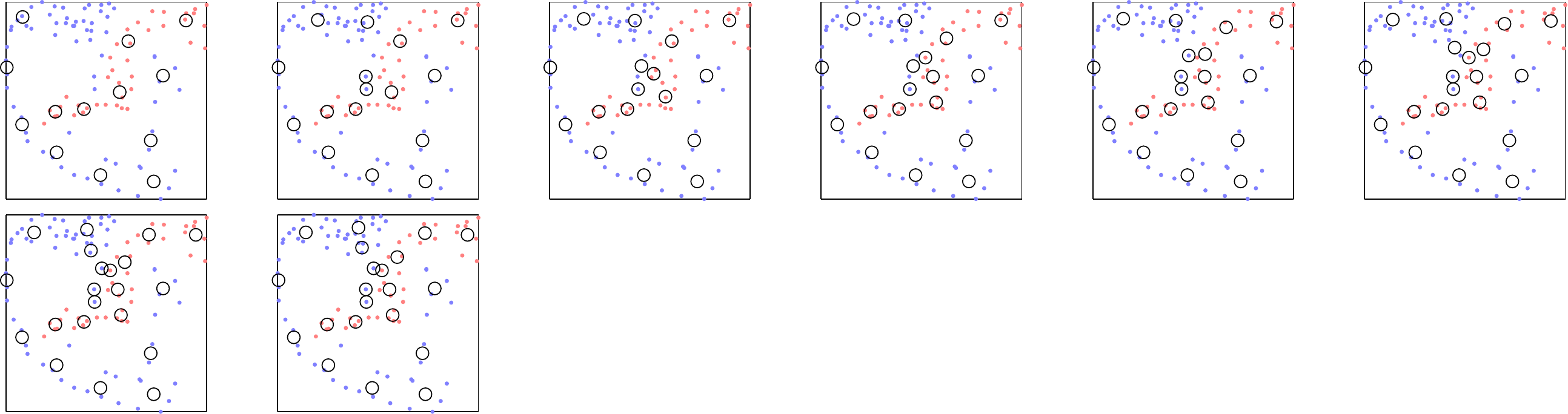}}\vspace{0pt}
  \caption{ The evaluation of the proposed method in three experiments on synthetic data. (a) The result on the first synthetic data set until convergence. (b) The result on the second synthetic data set (first set augmented with novel positively labelled data) until convergence. (c) The result on the second synthetic data set, initialized with the clustering result of the first experiment, until convergence. }
  \label{f:synth}
\end{figure*}

We start this section by illustrating the operation of the proposed algorithm on a synthetic 2D example. Consider the top-left plot in Figure~\ref{f:synth}(a). It shows positively (blue) and negatively (red) labelled data, constructed by hand. The subsequent plots in this figure (left to right, and then top to bottom) show the progress of the algorithm and the creation of new clusters, as old clusters with conflicting data member labels are split. We ran this experiment until convergence. Note the higher density of clusters where the two classes are closer and more sparsely distributed clusters where class membership is clearer.

Next, we modified the previous experiment by augmenting the positively labelled set with new data points, as can be seen in the top-left plot in Figure~\ref{f:synth}(b). As before, we re-run our algorithm from the start and until convergence, the subsequent plots in this figure (left to right, and then top to bottom) showing the progress. Compare the final, converged state of this experiment with that of the previous. In the regions of the plane distant from the region in which the new data was added, the revealed data structure i.e.\ the loci of the cluster centres, for both positively and negatively labelled data, remains the same. On the other hand, significant changes can be observed in the regions in which the two classes exhibit new proximity. Consequently, new cluster centres, densest where the two classes are the closest, are created to ascertain good discriminative performance.

Our last synthetic experiment was inspired by the observed localization of effects that the addition of new data has on our algorithm. In this experiment, we used the same as in the previous experiment, but instead of running it from the start as previously, we proceed from the clustering determined in our first experiment i.e.\ before new data was added. The progress of our algorithm is shown in Figure~\ref{f:synth}(c). Note that the locations of clusters in the top-left plot of Figure~\ref{f:synth}(c) is the same as in the bottom-right plot of Figure~\ref{f:synth}(a). As before, we proceeded until convergence. A comparison of the final results of this and the previous experiments reveals a remarkable agreement between the determined cluster centres loci.

\subsection{Face recognition}\label{ss:fr}
\begin{figure*}[ht]
  \centering
  \begin{tabular}{c}
    \includegraphics[width=0.104\textwidth]{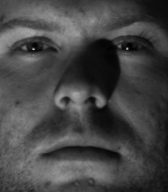}~~~\includegraphics[width=0.104\textwidth]{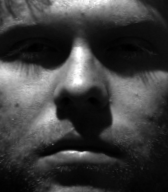}~~~\includegraphics[width=0.104\textwidth]{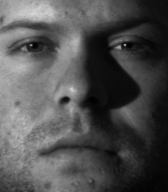}~~~\includegraphics[width=0.104\textwidth]{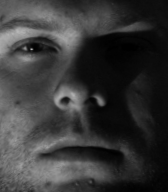}~~~\includegraphics[width=0.104\textwidth]{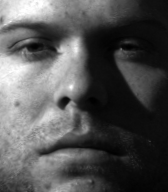}~~~\includegraphics[width=0.104\textwidth]{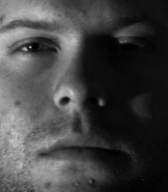}~~~\includegraphics[width=0.104\textwidth]{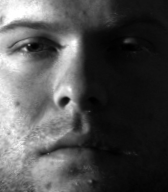}~~~\includegraphics[width=0.104\textwidth]{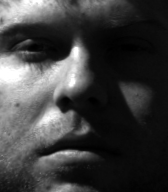}\vspace{5pt}\\
    \includegraphics[width=0.104\textwidth]{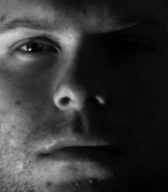}~~~\includegraphics[width=0.104\textwidth]{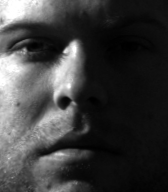}~~~\includegraphics[width=0.104\textwidth]{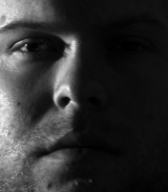}~~~\includegraphics[width=0.104\textwidth]{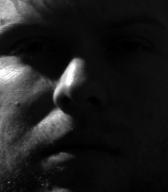}~~~\includegraphics[width=0.104\textwidth]{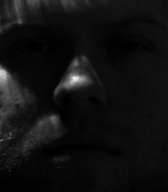}~~~\includegraphics[width=0.104\textwidth]{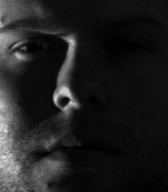}~~~\includegraphics[width=0.104\textwidth]{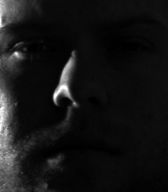}~~~\includegraphics[width=0.104\textwidth]{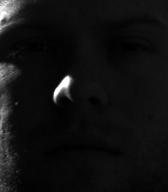}\vspace{5pt}\\
    \includegraphics[width=0.104\textwidth]{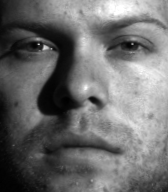}~~~\includegraphics[width=0.104\textwidth]{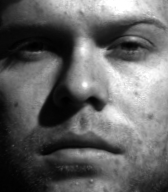}~~~\includegraphics[width=0.104\textwidth]{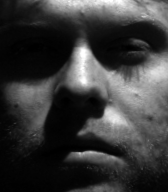}~~~\includegraphics[width=0.104\textwidth]{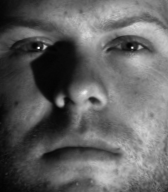}~~~\includegraphics[width=0.104\textwidth]{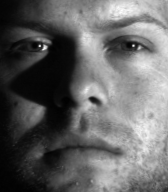}~~~\includegraphics[width=0.104\textwidth]{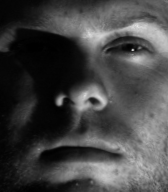}~~~\includegraphics[width=0.104\textwidth]{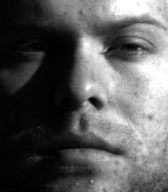}~~~\includegraphics[width=0.104\textwidth]{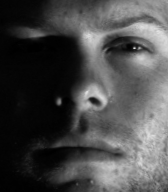}\vspace{5pt}\\
    \includegraphics[width=0.104\textwidth]{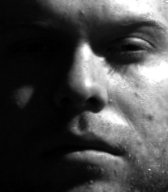}~~~\includegraphics[width=0.104\textwidth]{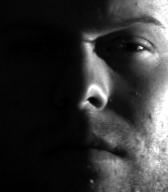}~~~\includegraphics[width=0.104\textwidth]{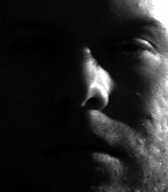}~~~\includegraphics[width=0.104\textwidth]{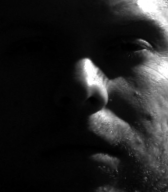}~~~\includegraphics[width=0.104\textwidth]{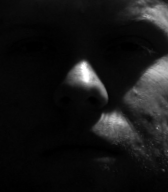}~~~\includegraphics[width=0.104\textwidth]{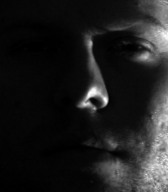}~~~\includegraphics[width=0.104\textwidth]{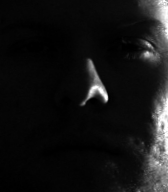}~~~\includegraphics[width=0.104\textwidth]{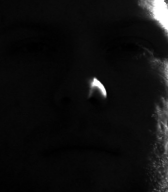}\vspace{5pt}\\
  \end{tabular}
  \caption{ Examples of extreme illuminations present in the data set used for the evaluation of methods in this paper.  }
  \label{f:yale}
\end{figure*}

Next, we evaluated the proposed algorithm on a challenging problem of immense practical significance: face recognition across illumination. As particularly appropriate for this experiment, we used the Extended YaleB database. This is a most difficult data set used as a standard benchmark for the comparison of face recognition algorithms in terms of their robustness to severe illumination changes. It contains 40 people and 64 images per person, each image corresponding to a different illumination. The variation in the direction of the dominant light source illuminating a face is extreme: its azimuth varies from -130$^\circ$ to 130$^\circ$, and its elevation from -40$^\circ$ (i.e.\ pointing upwards) to 90$^\circ$ (i.e.\ directly overhead, pointing downwards), giving a total of 64 different illumination conditions. Notice that the face is sometimes illuminated from the rear lateral direction (and thus hardly illuminated at all), that extreme cast shadows are often present as are highly bright saturated image regions. These challenges are illustrated in Figure~\ref{f:yale}. The database does not include any intentional variation in facial expression, but some variation exists nonetheless, mainly in the form of squinting when the subject is facing the dominant light source.

In this experiment, we adopted the leave-one-out approach. Specifically, we select an image of a face which will be used as a novel query face for classification. All other faces are used for training. After clustering is applied (the conventional $k$-means algorithm or the algorithm proposed herein), the query face is recognized as the person with the closest cluster centre (\textit{maximum maximorum}). We iterate through all the 2560 available images and use each of them as the novel face in turn. For the sake of a fair comparison, both for the conventional $k$-means algorithm and the discriminative $k$-means we used the parameter value of $k=8$ (in other words, for the proposed algorithm we would stop further refinement and terminate the algorithm when this target number of clusters is reached).

\begin{figure*}[thb]
  \centering
  \subfigure[Conventional $k$-means]{\includegraphics[width=0.45\textwidth]{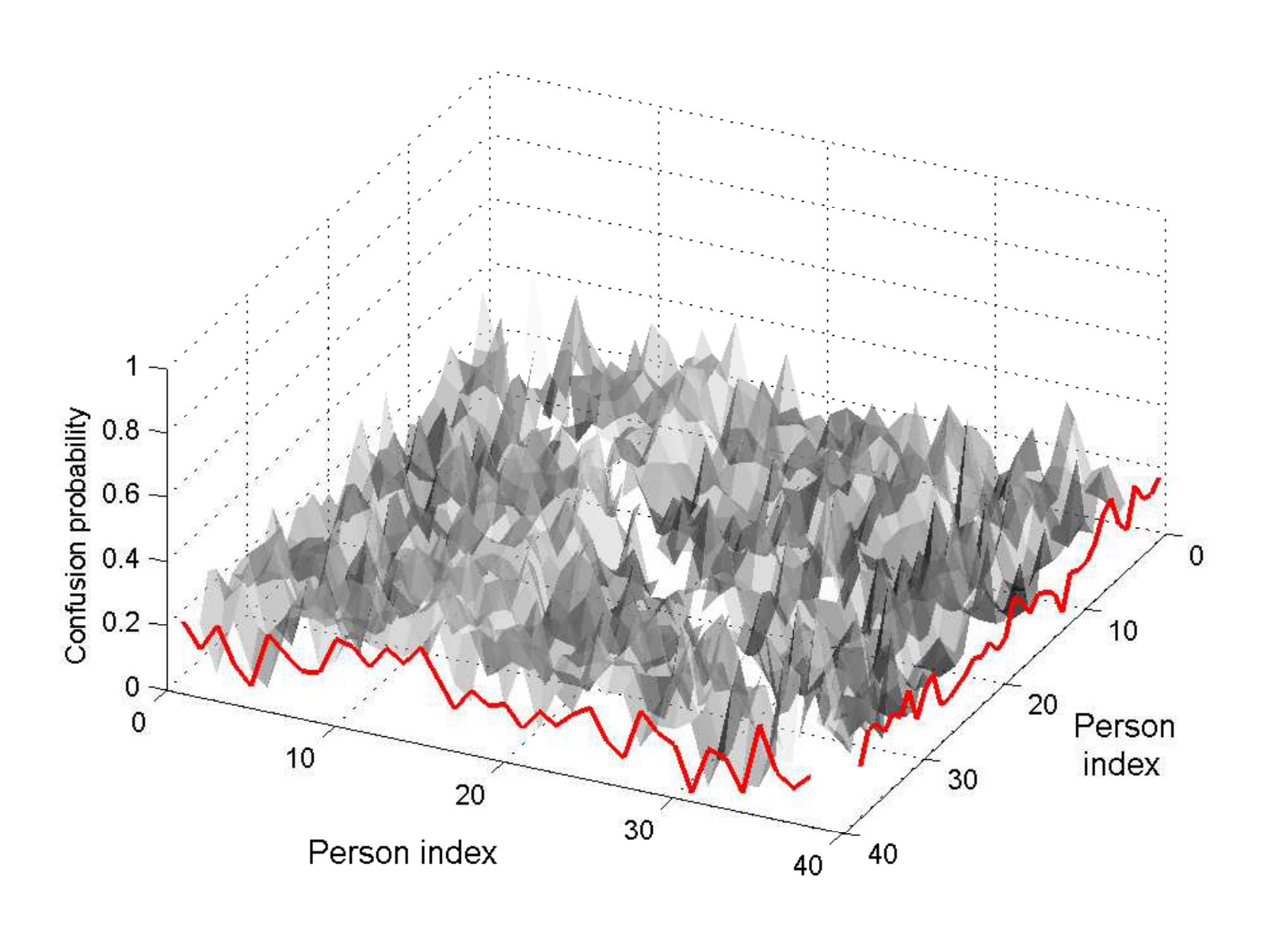}}
  \subfigure[Proposed discriminative $k$-means]{\includegraphics[width=0.45\textwidth]{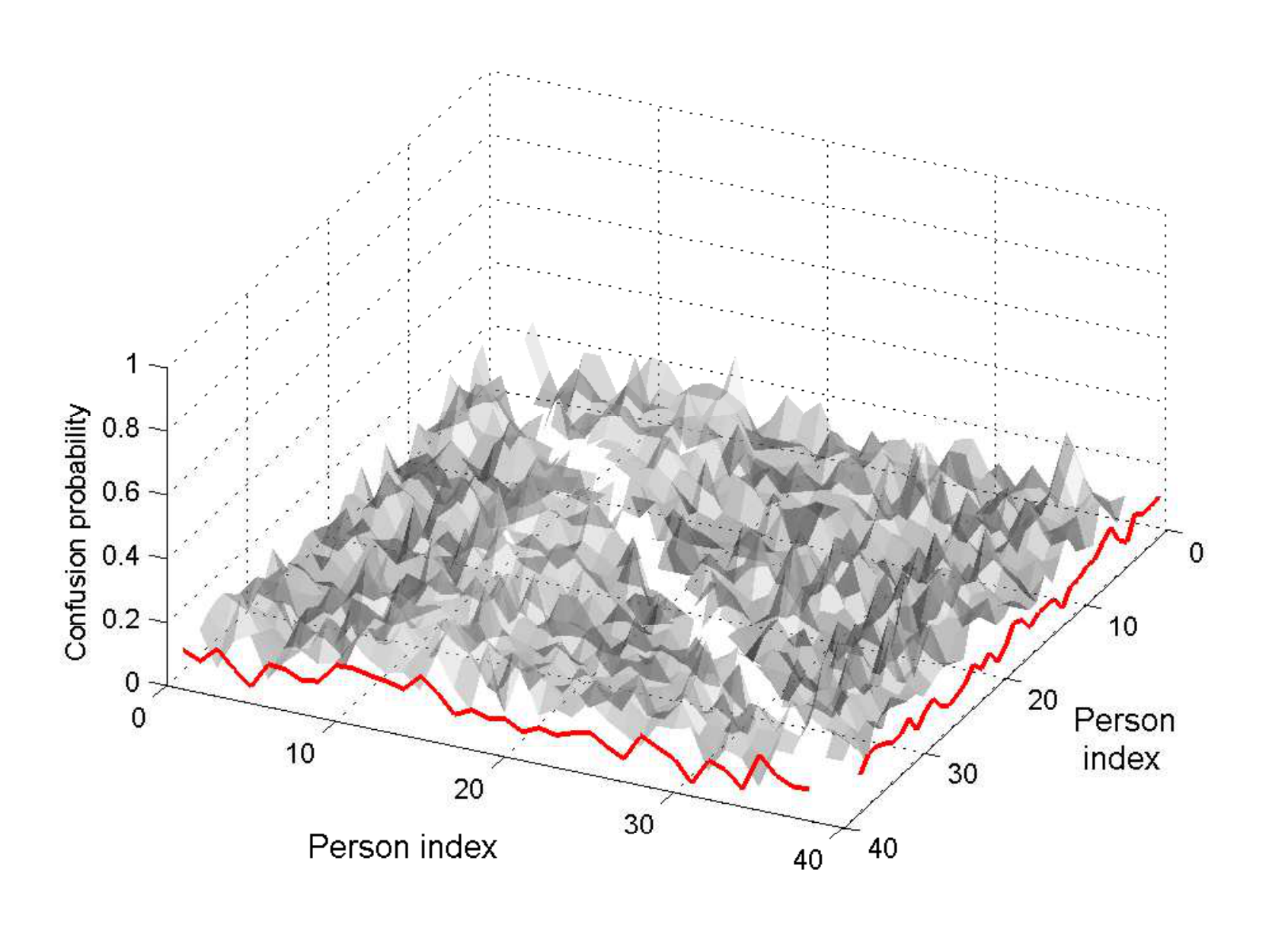}}
  \vspace{0pt}
  \caption{ Confusion matrices obtained on the Extended YaleB data set (a) when the conventional $k$-means algorithm was used to extract the cluster structure of each person's appearance, and (b) when the proposed discriminative $k$-means clustering was used instead. The 1D plots (red lines) show `marginalized' confusions. Note that the performance of the proposed algorithm is consistently superior, the error rate attained with the use of the conventional $k$-means algorithm being approximately 50\% higher throughout the data set. }
  \label{f:confREF}
\end{figure*}

The confusion matrix we obtained by using the conventional $k$-means algorithm in this evaluation framework is shown in Fig.\ref{f:confREF}(a). This figure shows the proportion of faces of one individual which were misclassified as another, for all pairs of individuals in the data set. In addition, the red lines show `marginalized' confusion, i.e.\ the proportion of faces of one individual misclassified as any other. The misclassification rate can be seen to be very high across the data set, with the average of 18\%. As expected from previous work, both in neurophysiology and computer-based face recognition, some individuals were more problematic than others as witnessed by the variations in the `marginalized' confusion.

Now, compare the error rate obtained by using the conventional $k$-means algorithm in Fig.\ref{f:confREF}(a) with that when the proposed discriminative $k$-means clustering is used instead, shown in Fig.\ref{f:confREF}(b). Our algorithm consistently achieved superior performance, resulting in the average error rate of approximately 12\%. At this point it should be repeated that this improvement is achieved under the constraint of the same number of target clusters -- the discriminative performance of the proposed method would have been improved further had the iteration been allow to proceed until convergence i.e.\ until the optimal number of clusters is reached; the same could not necessarily be expected with the conventional $k$-means algorithm.

\begin{figure}[thb]
  \centering
  \vspace{-20pt}
  \includegraphics[width=0.45\textwidth]{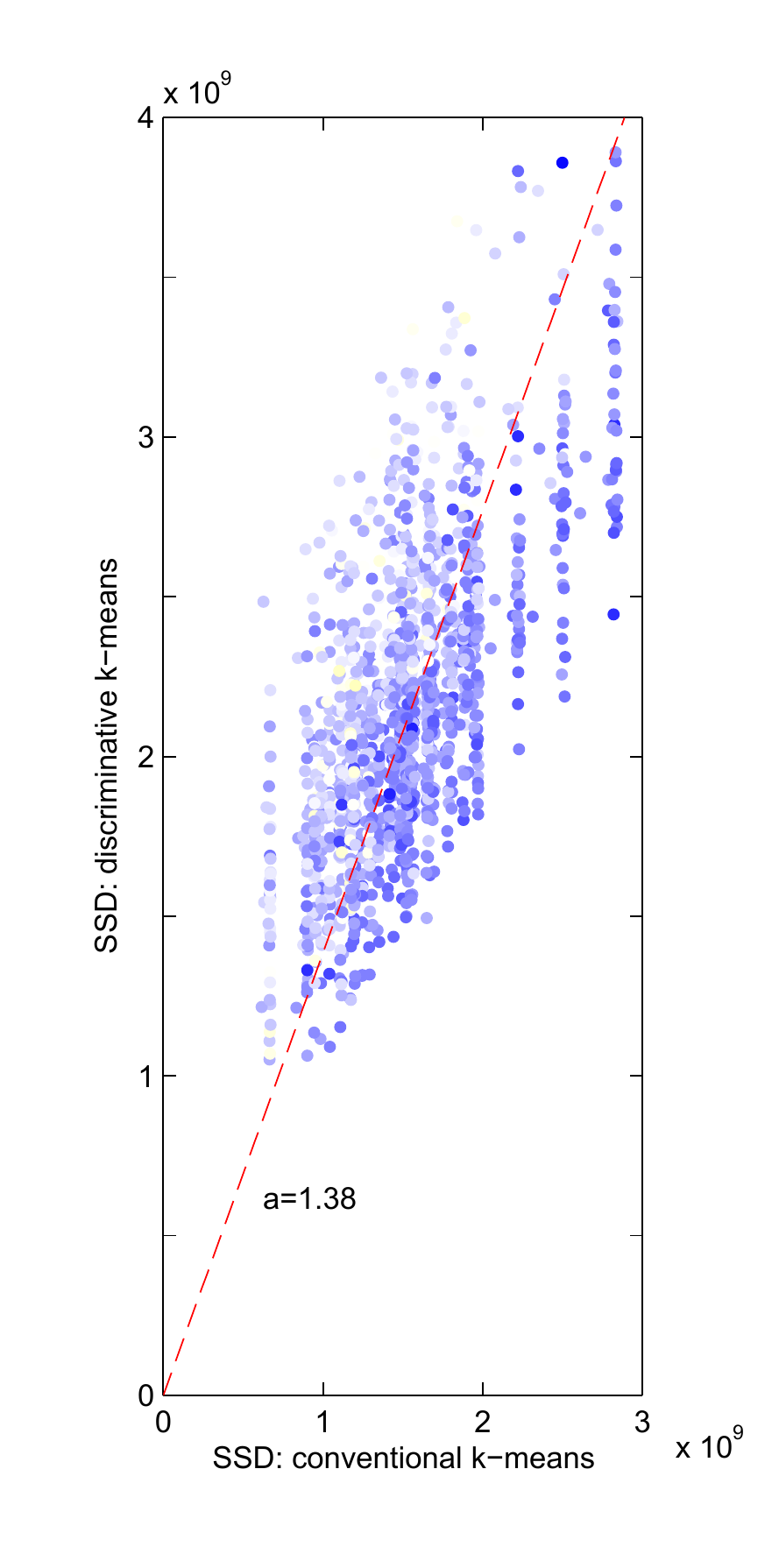}
  \vspace{-20pt}
  \caption{ The sum of squared distances from all data points to their corresponding cluster centres of the proposed method against that of the conventional $k$-means algorithm. Each circle corresponds to a particular face image being used as a leave-one-out novel query, as described in Section~\ref{ss:fr}. The `blueness' of the circle encodes the benefit of the proposed method over the conventional $k$-means algorithm -- the more intensely blue the circle is, the greater the advantage. }
  \label{f:SSD}
\end{figure}

Considering the superior performance of the proposed algorithm on the one hand, and the difference in its approach in comparison with the conventional $k$-means algorithm (which may be succinctly described as `discrimination' vs.\ `description'), we next examined the sum of squared distances (SSD) from all data points to their corresponding cluster centres for the two algorithms. The result is summarized in the plot of Fig.\ref{f:SSD} which shows as a circle each different clustering instance corresponding to a different face image being used as a leave-one-out novel query. As expected, the SSD of the proposed algorithm is higher than of the conventional $k$-means -- the red line shows the best straight line fit passing through the origin, its slope being $a \approx 1.38$. Note that we also colour-coded each datum by the benefit that the proposed algorithm demonstrated over the conventional $k$-means (the more intensely blue the circle is, the greater the improvement in recognition), as we were interested in investigating whether there is any relationship between this benefit and the obtained SSD. As the plot serves to demonstrate, no such relationship was observed.

\begin{figure}[tb]
  \centering
  \footnotesize
  \includegraphics[width=0.45\textwidth]{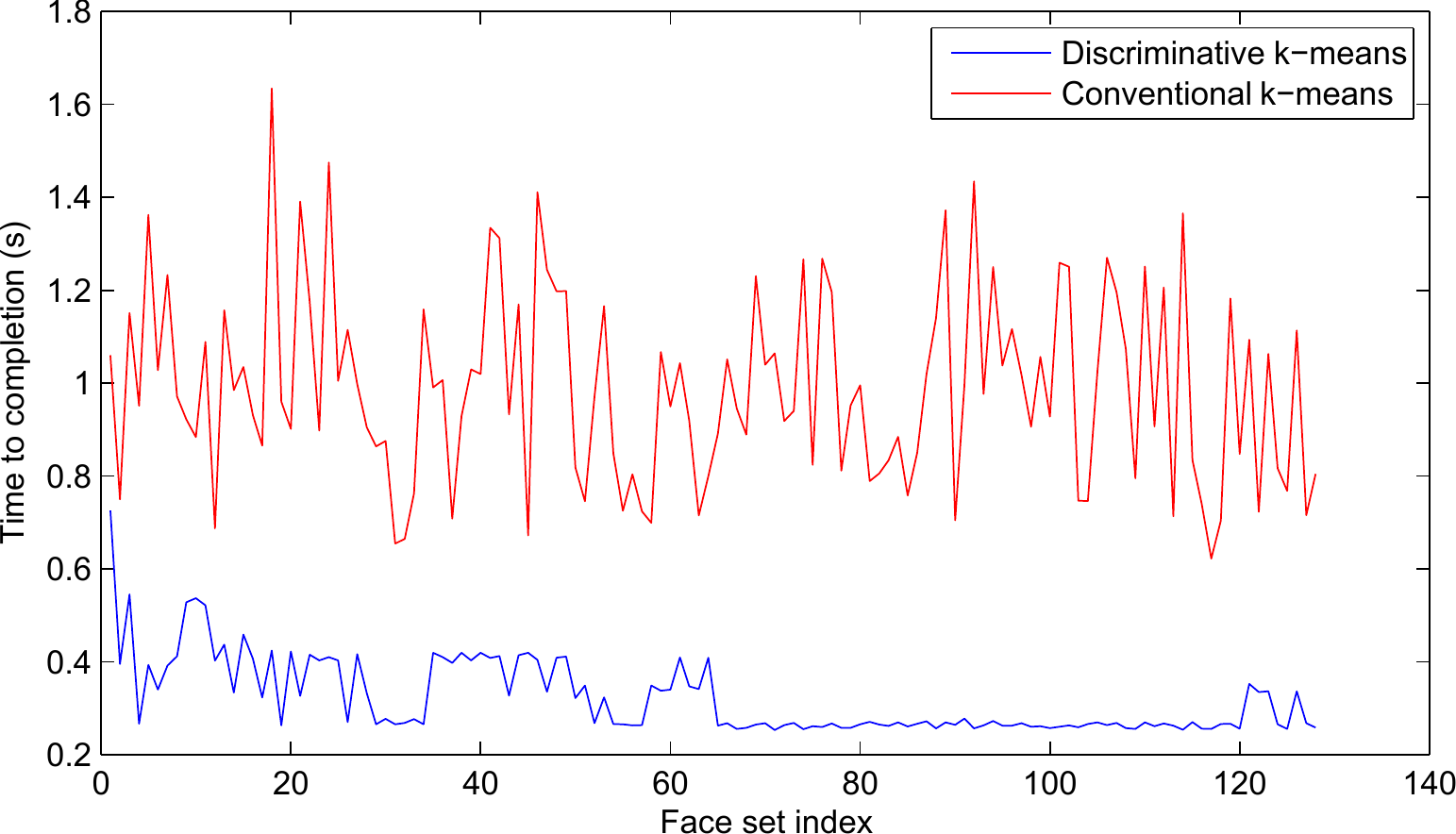}
  \caption{ Time until convergence for the $k$-means algorithm (red line) in the reported evaluation (the value of parameter was set to $k=8$), and the time until the desired number of clusters was reached by the proposed algorithm (blue line; as for the $k$-means algorithm, the parameter was set to $k=8$). }
  \label{f:timing}
\end{figure}

Lastly, we examined the running efficiency of the proposed method. The measured time until the present number of clusters $k=8$ was reached is plotted in Fig.\ref{f:timing} as a blue line, and compared with the time until convergence on the same data of the $k$-means algorithm. Perhaps surprisingly at first, the time required by the proposed algorithm is much shorter (by approximately 73\%). There are two key aspects of our method that explain this observation. Firstly, note that since our algorithm starts from a single cluster, most of the time the number of clusters that it handles (e.g.\ that distances from all points are evaluated to) is smaller than $k$. Secondly, unlike in the conventional $k$-means algorithm in which the cluster centres are initially randomly assigned, when new clusters are created in the proposed method, they are created in a purposeful manner and are by construction placed where new clusters are actually needed.

\section{Summary and conclusions}\label{s:conc}
In this paper we described a novel clustering algorithm inspired by the $k$-means algorithm. Unlike the $k$-means algorithm, the proposed method has the notion of two class labelled data, and aims to form clusters that best capture the structure of each class for discriminative purposes. Initialized with a single cluster, the proposed algorithm splits as needed and updates their centres using an equation which is a generalization of the update equation of the $k$-means algorithm. The superiority of our approach was demonstrated on both synthetic data and a real problem of face recognition across illumination changes.

\bibliographystyle{ieeetran}
\bibliography{./my_bibliography}

\begin{thebibliography}{10}
\providecommand{\url}[1]{#1}
\csname url@samestyle\endcsname
\providecommand{\newblock}{\relax}
\providecommand{\bibinfo}[2]{#2}
\providecommand{\BIBentrySTDinterwordspacing}{\spaceskip=0pt\relax}
\providecommand{\BIBentryALTinterwordstretchfactor}{4}
\providecommand{\BIBentryALTinterwordspacing}{\spaceskip=\fontdimen2\font plus
\BIBentryALTinterwordstretchfactor\fontdimen3\font minus
  \fontdimen4\font\relax}
\providecommand{\BIBforeignlanguage}[2]{{%
\expandafter\ifx\csname l@#1\endcsname\relax
\typeout{** WARNING: IEEEtran.bst: No hyphenation pattern has been}%
\typeout{** loaded for the language `#1'. Using the pattern for}%
\typeout{** the default language instead.}%
\else
\language=\csname l@#1\endcsname
\fi
#2}}
\providecommand{\BIBdecl}{\relax}
\BIBdecl

\bibitem{VoevBalcRoglTeng+2012}
K.~Voevodski, M.-F. Balcan, H.~R{\"{o}}glin, S.-H. Teng, and Y.~Xia, ``Active
  clustering of biological sequences.'' \emph{Journal of Machine Learning
  Research}, vol.~13, no.~9, pp. 203--225, 2012.

\bibitem{JangHend2007}
W.~Jang and M.~Hendry, ``Cluster analysis of massive datasets in astronomy.''
  \emph{Journal of Statistics and Computing}, vol.~17, no.~3, pp. 253--262,
  2007.

\bibitem{Lowe2004}
D.~G. Lowe, ``Distinctive image features from scale-invariant keypoints.''
  \emph{International Journal of Computer Vision}, vol.~60, no.~2, pp. 91--110,
  2003.

\bibitem{AranZiss2011}
R.~Arandjelovi{\'c} and A.~Zisserman, ``Smooth object retrieval using a bag of
  boundaries.'' \emph{In Proc. IEEE International Conference on Computer
  Vision}, pp. 375--382, 2011.

\bibitem{Aran2012f}
O.~Arandjelovi{\'c}, ``Object matching using boundary descriptors.'' \emph{In
  Proc. British Machine Vision Conference}, 2012, {DOI:~10.5244/C.26.85}.

\bibitem{JegoDouzSchmPere2010}
H.~J\'{e}gou, M.~Douze, C.~Schmid, and P.~P\'{e}rez, ``Aggregating local
  descriptors into a compact image representation.'' \emph{In Proc. IEEE
  Conference on Computer Vision and Pattern Recognition}, pp. 3304--3311, 2010.

\bibitem{Aran2011a}
O.~Arandjelovi{\'c}, ``Contextually learnt detection of unusual motion-based
  behaviour in crowded public spaces.'' \emph{In Proc. International Symposium
  on Computer and Information Sciences}, pp. 403--410, 2011.

\bibitem{AranCipo2006c}
O.~Arandjelovi{\'c} and R.~Cipolla, ``Automatic cast listing in feature-length
  films with anisotropic manifold space.'' \emph{In Proc. IEEE Conference on
  Computer Vision and Pattern Recognition}, vol.~2, pp. 1513--1520, 2006.

\bibitem{OchsBrox2012}
P.~Ochs and T.~Brox, ``Higher order motion models and spectral clustering.''
  \emph{In Proc. IEEE Conference on Computer Vision and Pattern Recognition},
  pp. 614--621, 2012.

\bibitem{MartAran2010}
R.~Martin and O.~Arandjelovi{\'c}, ``Multiple-object tracking in cluttered and
  crowded public spaces.'' \emph{In Proc. International Symposium on Visual
  Computing}, vol.~3, pp. 89--98, 2010.

\bibitem{Lian2013}
C.~Liang, ``Learning atomic human actions using variable-length {M}arkov
  models.'' \emph{IEEE Transactions on Multimedia}, 2013, (to appear).

\bibitem{GirgShipWilc2011}
A.~Girgensohn, F.~Shipman, and L.~Wilcox, ``Adaptive clustering and interactive
  visualizations to support the selection of video clips.'' \emph{In Proc.
  International Conference on Multimedia Retrieval)}, p.~34, 1997.

\bibitem{NgJordWeis2001}
A.~Y. Ng, M.~I. Jordan, and Y.~Weiss, ``On spectral clustering: analysis and an
  algorithm.'' \emph{Advances in Neural Information Processing Systems}, pp.
  849--856, 2001.

\bibitem{McCuYang2008}
P.~McCullagh and J.~Yang, ``How many clusters?'' \emph{Bayesian Analysis},
  vol.~3, no.~1, pp. 101--120, 2008.

\bibitem{RobeHolmDeni2001}
S.~J. Roberts, C.~Holmes, and D.~Denison, ``Minimum-entropy data clustering
  using reversible jump {M}arkov chain {M}onte {C}arlo.'' \emph{In Proc.
  International Conference on Artificial Neural Networks}, pp. 103--110, 2001.

\bibitem{DudaHartStor2001}
R.~O. Duda, P.~E. Hart, and D.~G. Stork, \emph{Pattern Classification},
  2nd~ed.\hskip 1em plus 0.5em minus 0.4em\relax New York: John Wily \& Sons,
  Inc., 2000.

\bibitem{Jain2010}
A.~K. Jain, ``Data clustering: 50 years beyond {k}-means.'' \emph{Pattern
  Recognition Letters}, vol.~31, no.~8, pp. 651--666, 2010.

\bibitem{Dasg2007}
S.~Dasgupta, ``The hardness of $k$-means clustering.'' University of
  California, San Diego, Technical Report CS2007-0890, 2007.

\bibitem{PenaLozaLarr1999}
J.~M. Pe{\~{n}}a, J.~A. Lozano, and P.~Larra{\~{n}}aga, ``An empirical
  comparison of four initialization methods for the $k$-means algorithm.''
  \emph{Pattern Recognition Letters}, vol.~20, no.~10, pp. 1027--1040, 1999.

\bibitem{KhanAhma2004}
S.~S. Khan and A.~Ahmadb, ``Cluster center initialization algorithm for
  $k$-means clustering.'' \emph{Pattern Recognition Letters}, vol.~25, no.~11,
  pp. 1293--1302, 2004.

\bibitem{Dunn1973}
J.~C. Dunn, ``A fuzzy relative of the {ISODATA} process and its use in
  detecting compact well-separated clusters.'' \emph{Journal of Cybernetics},
  vol.~3, pp. 32--–57, 1973.

\bibitem{KaufRous2005}
L.~Kaufman and P.~J. Rousseeuw, \emph{Finding groups in data : An introduction
  to cluster analysis.}\hskip 1em plus 0.5em minus 0.4em\relax Wiley, 2005.

\bibitem{SchoSmolMull1998}
B.~{Sch\"{o}lkopf}, A.~Smola, and K.-R. {M\"{u}ller}, ``Nonlinear component
  analysis as a kernel eigenvalue problem.'' \emph{Neural Computation},
  vol.~10, no.~5, pp. 1299--1319, 1998.

\end{thebibliography}

\end{document}